\documentclass{article}
\usepackage{ijcai23}

\usepackage{times}
\usepackage{soul}
\usepackage{url}
\usepackage[hidelinks]{hyperref}
\usepackage[utf8]{inputenc}
\usepackage[small]{caption}
\usepackage{graphicx}
\usepackage{amsmath}
\usepackage{amsthm}
\usepackage{booktabs}
\usepackage{algorithm}
\usepackage{algorithmic}

\usepackage[inline]{enumitem}

\usepackage{xspace}
\newcommand{\etal}{{et al.\@\xspace}}

\usepackage{amsfonts}

\usepackage{multirow}
\usepackage{array}

\usepackage{caption}
\usepackage{subcaption}

\newtheorem{example}{Example}

\title{What Lies beyond the Pareto Front?\\A Survey on Decision-Support Methods for Multi-Objective Optimization}

\author{
Zuzanna Osika$^1$\and
Jazmin Zatarain~Salazar$^1$\and
Diederik M.~Roijers$^{2,3}$\and\\
Frans A.~Oliehoek$^1$\And
Pradeep K.~Murukannaiah$^1$
\affiliations
$^1$ Delft University of Technology, The Netherlands\\
$^2$ Vrije Universiteit Brussel, Belgium\\
$^3$ City of Amsterdam, The Netherlands\\
\emails
\{z.osika, j.zatarainsalazar, f.a.oliehoek, p.k.murukannaiah\}@tudelft.nl,
diederik.roijers@vub.be
}

\begin{document}

\maketitle

\begin{abstract}
We present a review that unifies decision-support methods for exploring the solutions produced by multi-objective optimization (MOO) algorithms. As MOO is applied to solve diverse problems, approaches for analyzing the trade-offs offered by MOO algorithms are scattered across fields. We provide an overview of the advances on this topic, including methods for visualization, mining the solution set, and uncertainty exploration as well as emerging research directions, including interactivity, explainability, and ethics. We synthesize these methods drawing from different fields of research to build a unified approach, independent of the application. Our goals are to reduce the entry barrier for researchers and practitioners on using MOO algorithms and to provide novel research directions.
\end{abstract}

\section{Introduction}
\label{sec:introduction}

Most real-world applications include multiple stakeholders with diverse interests. Such problems are naturally formulated as multi-objective optimization (MOO) problems by representing the stakeholders' interests via objectives. The objectives correspond to the stakeholders' aims in an application, e.g., minimizing pollution, and should be operationalized using meaningful metrics, e.g., the density of fine particles in the air or the air quality index. Since the objectives may be conflicting, there may not be a single solution that is optimal for all objectives. Which solution should ultimately be selected depends on the people responsible for deciding which solution to execute. This could be a single person mandated to make this decision, a committee of stakeholders, or a political entity such as a city council. We will refer to these people as the  decision-makers (DMs).

The more complex the problem, the more unlikely that the DMs can express their preferences with respect to the objectives (even approximately) \textit{a priori}. As such, the DMs need to be informed about the available, possibly optimal, trade-offs. In MOO algorithms, it is common to produce a set of non-dominated solutions referred to as a Pareto-optimal set. A solution $x$ is (Pareto) non-dominated if there exists no other solution $y$ that is better than $x$ on at least one objective without being worse on any other objective. The set of all solutions non-dominated with respect to each other form the Pareto-optimal set. The projection of the Pareto(-optimal) set in the objective space is called the Pareto(-optimal) front. 

While the Pareto set is a general solution set, it might be excessive when more information about the DMs is available. Further, it can even be wrong when stochastic solutions are allowed but are not taken into account \cite{vamplew2009constructing}, or incomplete when the outcomes are stochastic and the DMs care about the expected utility for individual outcomes rather than having a utility for the expected outcome \cite{hayes2022expected}. Thus, we refer to the output of MOO as a \emph{solution set}, which can be but is not required to be a Pareto set.

Single-objective optimization (SOO) is often seen as an alternative to MOO. To employ SOO, important characteristics of the problem would need to be combined into a single, scalar, function. However, using SOO for a problem with many objectives has disadvantages \cite{practical-Hayes-2022}. First, finding a suitable combined-objective function (often, a manual process) is quite challenging and it may require simplifying assumptions, e.g., that the objectives are linear additive. Second, SOO is less adaptive to evolving objectives---adding or removing an objective requires re-engineering the objective function. Finally, combining multiple objectives into one function loses information, particularly, when it is not possible to collapse the underlying goals (e.g., the environmental and economic objectives, which have different unit values and levels of risk) into a single measure. As a result, an SOO solution is less informative to a DM than an appropriate solution set produced by MOO. For example, with an SOO solution, a DM can typically only know the solution's scalar objective value, but with a solution set, the DM can compare solutions in terms of the problem characteristics.

\begin{figure}[!htb]
    \centering
    \includegraphics[width=\columnwidth]{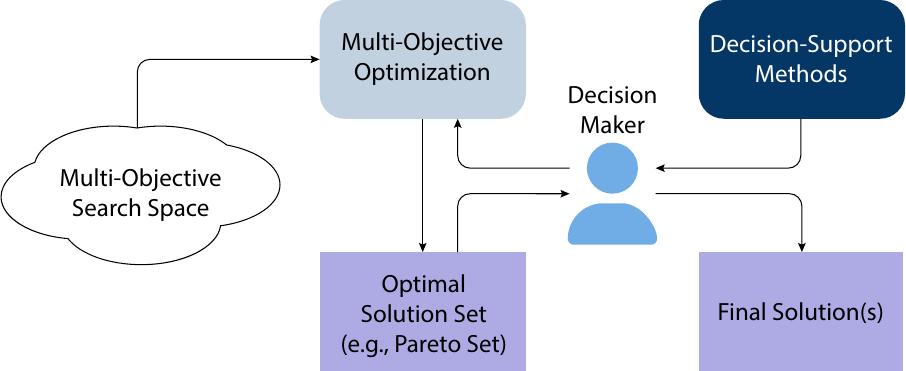}
    \caption{Multi-objective decision-making involves computational optimization as well as a DM's analysis of the solution set produced by the optimization algorithm. This survey focuses on the methods for supporting the DM in reaching the final solution(s).}
    \label{fig:moo-overview}
\end{figure}

As the number of objectives increases, the number of solutions in the solution set produced by an MOO algorithm is also likely to increase. For example, for a problem with five objectives, the size of the solution set can be in the order of hundreds. However, for most problems, only a few final solutions (often, only one) are desired. For example, if the problem is to find an optimal design for a car engine, the car manufacturer may only want one design to send to production. Thus, a DM must analyze the solution set outputted by MOO to identify the final solution as shown in Figure~\ref{fig:moo-overview}. 

Even with as few as three or four objectives, analyzing the trade-offs among the solutions can be overwhelming. Further, considering multiple decision variables, dependencies among these variables, and external factors (e.g., uncertainties) influencing the optimization makes the analysis even more complex. Then, how can a DM systematically explore the alternative solutions to produce the final solution(s)?

Unfortunately, there is no easy answer to this question. There is no standard procedure or standard set of methods a DM can adopt to explore the output of MOO. Further, as MOO is applied in diverse fields,  methods have been developed in different silos. Thus, there is a need to bring together these decision-support methods in a systematic manner.

We perform a comprehensive review of decision-support methods for MOO. Our review includes methods for 
\begin{enumerate*}[label=(\arabic*)]
    \item visualizing solution sets such as the Pareto front, 
    \item extracting the knowledge from the solution sets with data mining techniques, and
    \item exploring uncertainty.
\end{enumerate*}
Whereas these are established lines of work, there are several emerging directions, including interactive methods, explainability, and providing support on ethical aspects such as distributive justice. We also discuss these directions. The two (non-conflicting) objectives of our work are to provide novel directions for researchers and reduce the entry barrier on using MOO for practitioners.

\paragraph{Existing Surveys on MOO}
MOO is gaining increasing attention as the advances in computational capabilities enable the application of MOO to problems having increasingly large search spaces and number of objectives. Accordingly, there have been several surveys on MOO. For example:

\begin{itemize}
    \item Li {\etal} \shortcite{Many-Li-2015} survey techniques for many objective optimization (a term used for MOO with at least four objectives) and identify seven categories of techniques.
    \item Tian {\etal} \shortcite{Tian-2021-Evolutionary} survey evolutionary MOO techniques. Antonio and Coello Coello \shortcite{Coevolutionary-Miguel-2018} survey coevolutionary algorithms, which are an extension of traditional evolutionary algorithms, for large-scale MOO.
    \item Hayes {\etal} \shortcite{practical-Hayes-2022} survey multi-objective reinforcement learning and planning techniques, and argue for a utility-based approach where the appropriate solution set is derived from what is known about the problem and the DM's utility.
\end{itemize}

Surveys such as the ones above focus on the optimization methods. In contrast, we seek to review decision-support methods, a step following optimization (Figure~\ref{fig:moo-overview}), though these steps may be used iteratively during  decision-making.

There have been a few surveys on specific aspects of decision support for MOO. For example:
\begin{itemize}
    \item Bandaru {\etal} \shortcite{Bandaru-2017-ESA-DataMiningMOO} survey exploratory data mining methods for extracting knowledge from MOO output.
    \item Moallemi {\etal} \shortcite{Moallemi-2018-SMPT-MOOUncertainty} survey exploratory modeling methods for analyzing the robustness of MOO solutions under deep uncertainty.
    \item Wang {\etal} \shortcite{Wang-2017-CIS-MOOPreference} survey preference modeling methods to direct a decision-maker to a region of the Pareto front.
\end{itemize}

To the best of our knowledge, none of the existing surveys provide a comprehensive review of decision-support methods, including all the dimensions we cover in this survey.

\paragraph{Organization} Section~\ref{sec:moo-variants-examples} formulates an MOO problem, and introduces different variants of the problem. Section~\ref{sec:decision-support-methods} introduces the three established categories of decision-support methods we review. Section~\ref{sec:directions} includes emerging research directions. Section~\ref{sec:conclusions} concludes the paper.

\section{Problem Variants and Examples}
\label{sec:moo-variants-examples}
As different types of MOO problems require different exploration techniques, we begin with a brief overview of MOO problems and variants, and provide motivating examples.

\subsection{Multi-Objective Optimization}

An MOO problem with $K$ objectives, $f_k(x): k = 1,....,K$, involves optimizing (maximizing or minimizing) for all objectives, simultaneously. Typically, a solution $x \in \mathbb{R}^n$ is a vector of $n$ decision variables, $x = (x_1,\ldots,x_n)$, which can be subject to constraints. Each objective function maps a solution to an objective value. Thus, each solution can be mapped to a point, $z = (z_1,\ldots,z_K)$, in the objective space. Alternatively, as is common in modern sequential decision making (RL or planning), a solution is described as a mapping from states to a probability distribution over actions. While such a mapping can still be cast as a vector of decision variables when both the set of possible states and the set of actions are discrete, when either is infinite (or prohibitively large) this is is no longer possible and the mapping becomes e.g., a neural network. Further, it is also possible that outcomes are stochastic, and it may be useful or in fact necessary to communicate not a single (expected) value for $z$, but rather a distribution over possible outcome vectors, $P(z|x)$.

Since the objectives of an MOO problem can be conflicting with each other, the output of the optimization is typically a set of solutions. For example, a Pareto-optimal set \cite{Multi-Deb-2011} is the typical solution set produced by evolutionary MOO techniques. In contrast, if the solutions or outcomes can be stochastic, we may need to produce a stochastic mixture \cite{vamplew2009constructing} (for sequential decision making problems, stochastic selection of one of the deterministic base policies at the start of each episode) or a set of distributions over outcomes \cite{hayes2022expected} as the solution set (built to maximise the expected utility).

\subsection{Dimensions of the Problem}
\label{sec:dimensions_of_the_problem}

We identify two dimensions of MOO problems, which influence the type of decision support required.

\paragraph{Preference Availability}
Knowing the DM's preferences is a key aspect of multi-objective decision-making. Such preferences can be elicited
\begin{enumerate*}[label=(\arabic*)]
\item \emph{a priori}, i.e., before the optimization process,
\item \emph{a posteriori}, i.e., after the optimization process, or
\item \emph{interactively} during the optimization process.
\end{enumerate*}

When the preferences are known \textit{a priori}, combining different objectives into a single objective may be possible \cite{Castelletti-multiobjective-2013}, but not always desirable (e.g., scalarization of unknown utility function may result in too much uncertainty) \cite{Survey-Roijers-2013}. However, research on preference construction \cite{Warren-2011-WCS-PreferenceConstruction} suggests that preferences are context-sensitive, and are often calculated at the time a choice is to be made. Thus, \textit{a posteriori} or interactive elicitation of preferences is typical. Such scenarios require decision support as the volume and the complexity of the choices MOO provides \cite{Zintgraf-2018-AAMAS-MODMPreference} can be overwhelming.

\paragraph{Solution Type}
The type of solution produced by MOO may call for different types of decision support. In simple cases, the solutions are one shot, e.g., MOO yields an optimal design for an engine that is put into production. In contrast, in complex problems the solution can consist of decision variables that need to be implemented over time (e.g., over several years) or space (e.g., across several countries). Such problems require decision support for analyzing, e.g., the time sensitivity \cite{What-Quinn-2019} of the solutions.

\subsection{Uncertainty Handling}
\label{sec:moo-variants-uncertainty}

In complex decision-making problems, such as policy and planning problems, uncertainty about the future has to be considered.  The main goal for uncertainty exploration in MOO is to help the DMs in making informed decisions by providing them with a comprehensive understanding of the range of possible solutions and their associated uncertainty. 

There are a number of methods to understand uncertainty; in this paper, we focus on stochastic uncertainty and deep uncertainty \cite{Kwakkel2016CopingUncertainty}. Stochastic uncertainty can be represented by a probabilistic model of random phenomena. Random variables are central to stochastic models. They often refer to natural phenomena,  for instance, next year’s rainfall or next month’s water consumption patterns. If we obtain several observations of that variable, we can estimate its probability distribution along with various statistical measures that characterize its distribution.  This style of uncertainty is often represented within the simulation which is coupled with MOO with which we can then obtain a probability distribution of the outcomes. In contrast, deep uncertainty \cite{Lempert-Robust-2019} refers to the uncertainty in the system that does not have a probabilistic representation due to the lack of observations. Deep uncertainty is concerned with variables whose statistical behavior is unknown. This concept has gained traction for recent decision support applications \cite{popper2019robust}, but is not new \cite{bertsekas1971recursive}.

The type of decision support depends on the nature of uncertainty, its location in the model, and severity. The decision-support methods also depend on whether uncertainty is analyzed after \cite{Impact-McPhail-2020} or during the optimization \cite{considering-Bartholomew-2020}.

\subsection{Motivating Examples}
\label{sec:motivating_examples}
Table~\ref{tab:moo-variants-examples} shows sample problems, chosen from different domains, for MOO variants. We also present a sample problem \cite{Yasin-2022-MSThesis-MOONileCase} in detail to illustrate the problem dimensions.

\begin{table}[!htb]
    \centering
    \begin{tabular}{@{}l>{\raggedright\arraybackslash}p{4.5cm}@{}}
    \toprule
     MOO Variant & Example problem and reference \\
     \midrule
     A posteriori preference & Combined heat and power generation \cite{two-Li-2018} \\
     Interactive  preference & Finnish forest management \cite{Towards-Misitano-2022}\\
     Uncertainty handling & Production allocation problem \cite{Visualizations-Babooshka-2021} \\
     One-shot solution & Building performance \cite{Three-Ling-2018}\\
     Sequential solution & Multireservoir operating policies \cite{What-Quinn-2019}\\
     \bottomrule
    \end{tabular}
    \caption{Example problems for MOO variants.}
    \label{tab:moo-variants-examples}
\end{table}

\begin{example}[Reservoir management]
\label{example:reservior}
The Nile Basin, which covers ten countries, is a crucial resource for supplying water for hydropower generation, municipal, industrial and agricultural consumption. However, tensions have risen between Ethiopia (upstream country), and Egypt and Sudan (downstream countries) over Ethiopia's construction of the Grand Ethiopian Renaissance (GERD) dam that could block the flow of water to downstream countries and threaten their water security. Thus, it is crucial to agree on the water release policy for the four reservoirs (one in Egypt, two in Sudan, and one in Ethiopia) for optimal water management. This is an MOO problem with conflicting objectives such as minimizing water demand deficit in Egypt and Sudan, maximizing hydro energy generation in Egypt and Ethiopia. The problem involves hydro-climatic and socio-economic uncertainties, as well as uncertainties regarding yearly water demand growth, and hydrology of the major tributaries of the river. A solution is a sequence of release decisions over the four reservoirs, at the beginning of each month, over a 20-year time horizon.
\end{example}

\section{Decision-Support Methods}
\label{sec:decision-support-methods}

Decision-support methods for MOO are often developed in a problem-specific manner. Yet, these methods have common building blocks. We review three categories of decision-support that are well-studied in the literature.

\subsection{Visualization}
\label{sec:visualization}

Visualizations are a common decision-support tool for exploring an MOO solution set. Miettinen \shortcite{Survey-Miettinen-2014} surveys graphical methods, e.g., bar charts, value paths, spider web charts, for visualizing a small set of alternatives in a solution set. However, as the number of objectives, and consequently, the number of alternative solutions increases, visualizing the solution set becomes extremely difficult. 

\begin{figure}[!htb]
    \centering
    \begin{subfigure}{\columnwidth}
        \includegraphics[width=\columnwidth]{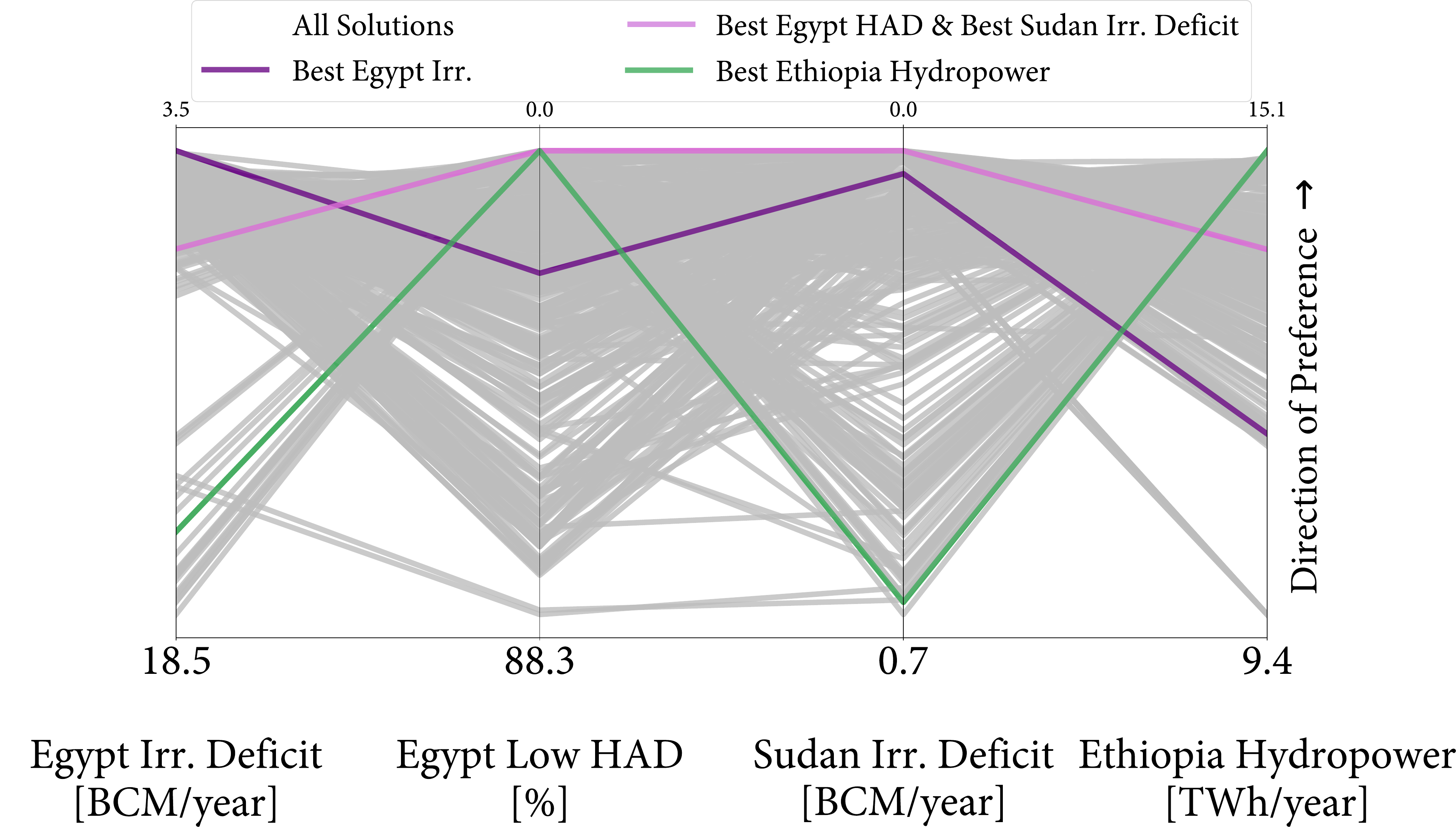}
        \label{subfig:PCP}
        \caption{A parallel coordinates plot, representing $N$-dimensional data by $N$ equally spaced, parallel, axes. The polylines represent solutions and they bisect each axes based on their values for objectives.}
    \end{subfigure}
    \begin{subfigure}{\columnwidth}
        \includegraphics[width=\columnwidth]{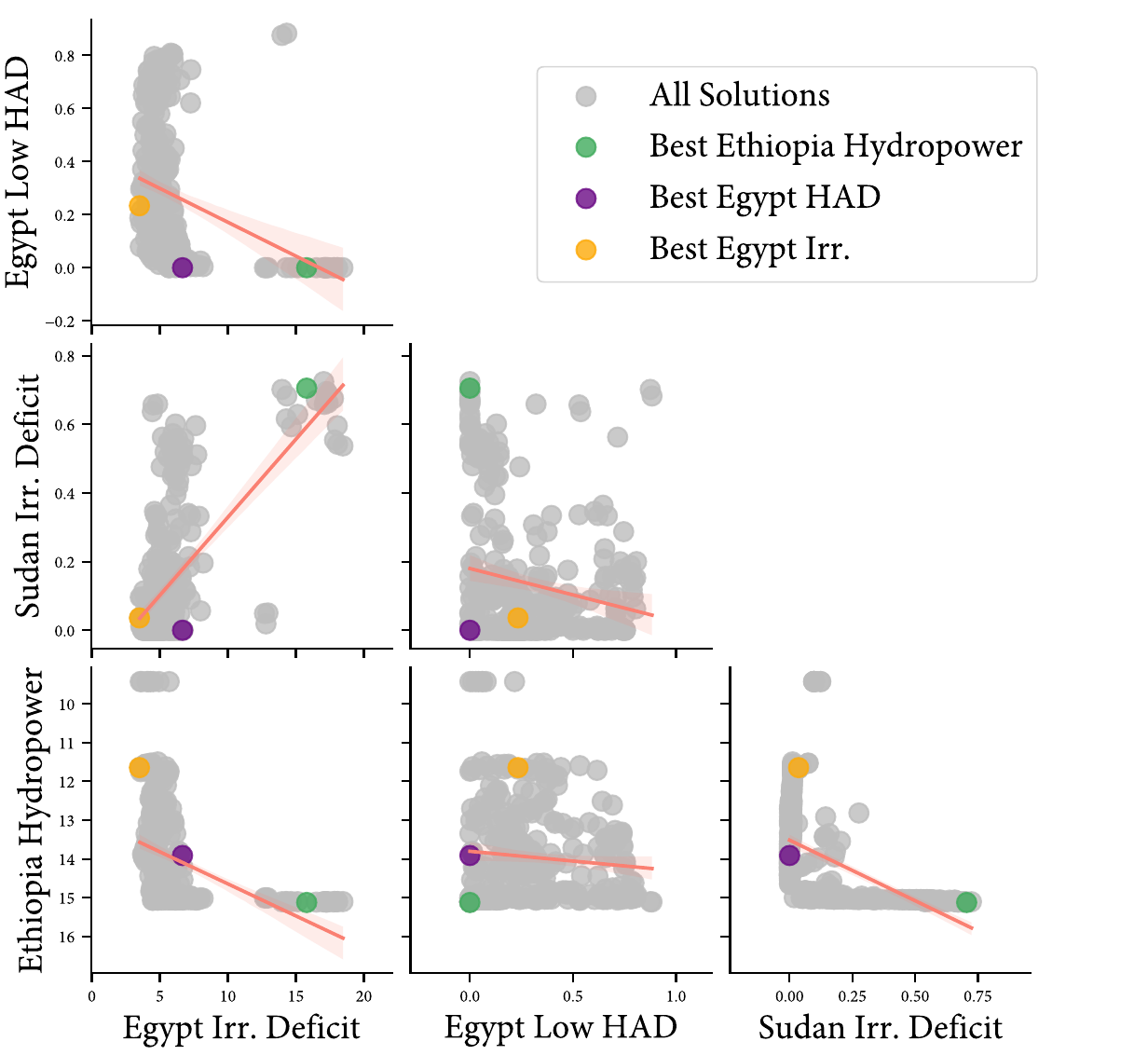}
        \label{subfig:scatter-plot}
        \caption{Pair-wise scatter plots, comparing solutions for each pair of objectives and indicating the general trend with regression lines.}
    \end{subfigure}
    \caption{Sample visualizations of the Pareto-optimal solutions to a simplified version of reservoir management problem (Example~\ref{example:reservior}).}
    \label{fig:moo-visuals}
\end{figure}

For problems with many objectives, the common visualizations employed include parallel coordinate plots (PCPs), pair-wise scatter plots, heat maps, and radar charts \cite{Improving-Dy-2022}. The PCP gives a comprehensive overview of all the solutions, and the other plots assist in further analyzing specific solutions, objectives, or their combinations. For instance, Figure~\ref{fig:moo-visuals} shows example plots for a simplified (four objectives) version of the reservoir management problem in Example~\ref{example:reservior}. As shown, the PCP looks quite cluttered with a large number of solutions. Specific solutions, e.g., best solutions for each objective, can be highlighted in the PCP. However, the extreme solutions may not be the most suitable solutions. The number of plots in the pairwise scatter plots increases combinatorially with the number of objectives. Then, tracking how specific subset of solutions fare across different pairs of objectives becomes quite challenging.

Several software tools have been developed, across application domains, for visualizing the solutions, e.g., PAVED \cite{PAVED-Cibulski-2020} (a web app), Parasol \cite{Parasol-Raseman-2019} (a Javascript library), and EMA Workbench \cite{Exploratory-Kwakkel-2017} (a Python library). Despite overlapping features, most of the existing visualization tools are developed independently---they don't build on a common core and only offer static visualizations \cite{Parasol-Raseman-2019}. 

Since visualizing a many-dimensional solution set is challenging, the dimensions of the solutions can be reduced (typically, to 2D or 3D). For instance, Nagar {\etal} \shortcite{Interpretable-Nagar-2021} propose to use interpretable self-organizing maps (iSOM), which works similar to a conventional SOMs in mapping a high-dimensional space to a low-dimensional space but differ in the way the \emph{best matching unit} is chosen in order to reduce folds and intersections in the low-dimensional space. Elewah {\etal} \shortcite{Elewah-2021-CEC-3DRadViz} propose 3D radial coordinate visualization (3D-RadViz), which maps a many dimensional objective space into 3D, preserving some properties of the solution set. However, since mapping a solution set to a low-dimensional space typically involves non-linear transformations, preserving the exact geometry of the solution set is not possible.

In contrast to works that visualize the MOO \emph{output}, Walter {\etal} \shortcite{Walter-2022-GECCO-ExplainableSearch} propose Population Dynamics Plot (PopDP) to visualize the MOO \emph{process} (specifically, for evolutionary MOO). PopDP shows not only the solutions in the objective space, but also the parent-offspring relationships and the perturbation operators that yield the solutions to show how the MOO solutions evolve through iterations.

\subsection{Mining the Solution Set}
\label{sec:mining_the_solution_set}
Visualizing a multi-dimensional solution set, in its entirety, is cognitively difficult. Although a visualization can present complex information, as Dy {\etal} \shortcite{Improving-Dy-2022} find there is a `ceiling' on the number of dimensions a DM can consider simultaneously. Thus, data mining methods---both supervised and unsupervised---have been developed to extract targeted information from a solution set to augment the high-level insights from visualizations. To apply these methods, we need to build an \emph{MOO dataset} consisting of input and/or output features from an MOO solution set.

In supervised methods, the input features are typically derived from the decision variables and the output feature from, e.g.,
\begin{enumerate*}[label=(\arabic*)]
\item ranks obtained by non-dominated sorting solutions;
\item one of the objective functions; 
\item preference information elicited from the decision-maker; or 
\item clustering methods \cite{Bandaru-2017-ESA-DataMiningMOO}.
\end{enumerate*}
Since the goal of such methods is to extract knowledge in a human-perceivable way, black-box models such as neural networks are typically not used. In contrast, methods such decision trees and logistic regression, which are easier to interpret, are used. For instance, Dudas {\etal} \shortcite{Post-Dudas-2015} use decision trees for the post-analysis of MOO solutions by utilizing the whole set of feasible solutions to find rules separating preferred from undesirable solutions. 

Unsupervised methods do not require `labeling' a feature of an MOO dataset as the output feature. For instance, a variety of methods have been applied to cluster the solution set in the objective space \cite{Bandaru-2017-ESA-DataMiningMOO}. Ulrich \shortcite{Ulrich-2013-JMCDA-BiobjectiveClustering} describes a method to find clusters that are compact and well-separated in both objective and decision spaces (multi-dimensional spaces defined by the objective functions and decision variables). Since good clusters in decision space may not correspond to good clusters in the objective space, Ulrich models clustering as a biobjective optimization problem. Sato {\etal} \shortcite{Sato-2019-ESA-MOTOAssociationRules} apply clustering and association rule mining (another unsupervised method) in sequence, where clustering groups solutions and association rules within a cluster provide finer insights. Bandaru \shortcite{Bandaru-2013-PhDThesis-AutomatedInnovization} develops an automated \emph{innovization} (innovation through optimization) approach, which discovers design principles that relate various problem elements (e.g., decision variables, objectives, and constraint functions) in an automated manner, employing grid-based clustering and genetic programming.

\subsection{Uncertainty Exploration}
\label{sec:uncertainty_exploration}

\begin{figure}[!htb]
    \centering
    \includegraphics[width=\columnwidth]{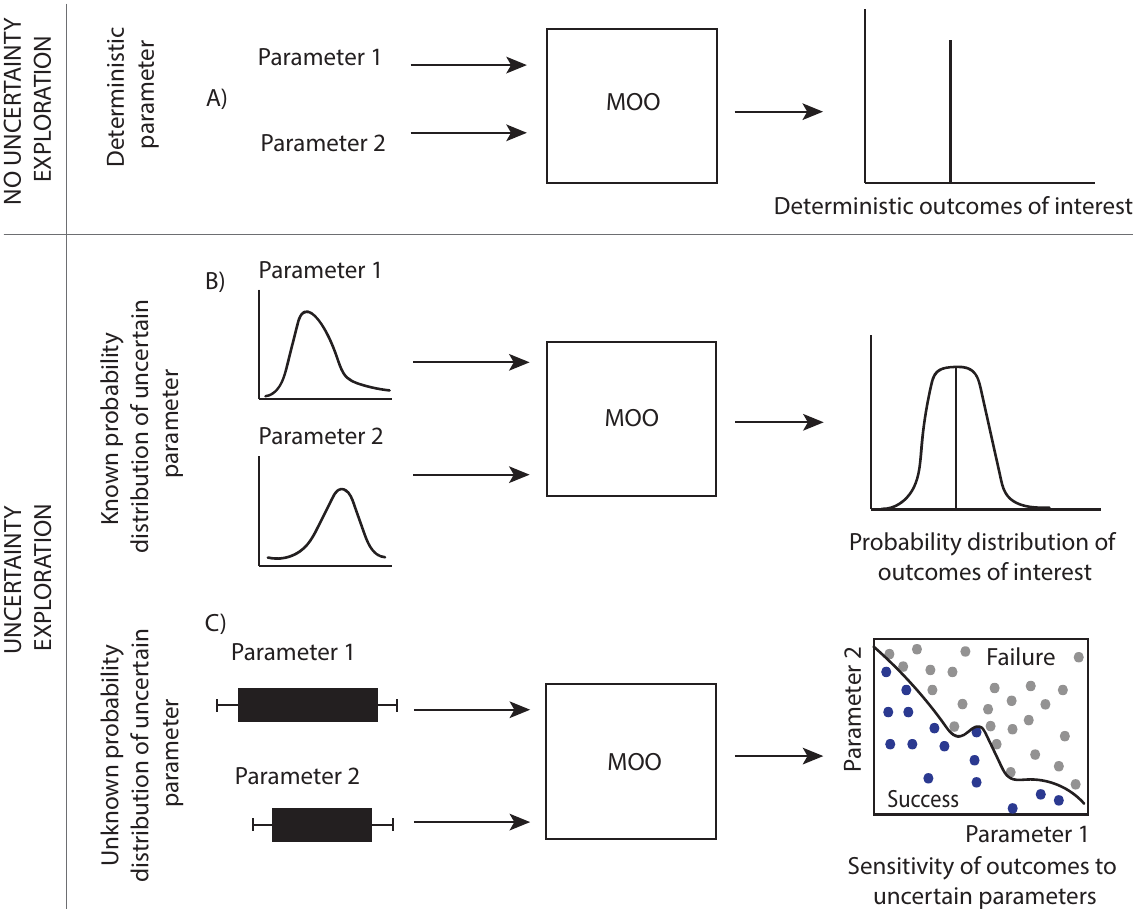}
    \caption{Uncertainty exploration within MOO. Panel A depicts no uncertainty exploration. Panel B shows uncertainty exploration by sampling the uncertain parameters and solving them multiple times to obtain a range of outcomes. Panel C shows a range of possible values of the uncertain parameters, which can be explored within the optimization or after the outcomes are known, e.g., to find the ranges in which the  outcomes of interest succeed (or fail) to meet a performance threshold.}
    \label{fig_levels_of_uncertainty}
\end{figure}

As mentioned in Section~\ref{sec:moo-variants-uncertainty}, we focus on the exploration of stochastic and deep uncertainty within MOO, which can be done within the optimization or at a post-processing stage. Uncertainty exploration within optimization involves exploration of solutions under a large ensemble of scenarios yielding optimization results that may perform well under a broader set of challenging scenarios. This style of exploration is referred to as multi-objective robust optimization \cite{shavazipour2021multi}. In contrast, in the post-optimization exploration, the goal is to find the combination of uncertain parameters and their ranges that can impact the outcomes of interest. For instance, the combinations and/or ranges of uncertain parameters that fail or succeed to meet given performance thresholds across objectives can be quantified. This style of exploration is particularly relevant for long-term planning.  Figure~\ref{fig_levels_of_uncertainty} showcases different approaches for uncertainty exploration within MOO (adapted from \protect\cite{salazar2022multi}).

With the vast climatic, technological, economic and socio-political changes it is no longer possible to determine how the future conditions might change, especially when considering long-term planning horizons (e.g., on the order of 70–100 years).  Nonetheless, decisions are still made under these conditions, with the additional difficulty that different stakeholders cannot agree upon, or do not have enough knowledge about how important are the various outcomes of interest; what are the relevant exogenous inputs to the system, and how they will change in the future \cite{Classifying-Kwakkel-2010}.  A number of techniques have been developed to cope with the challenges of decision-making under uncertainty, particularly when the decisions taken today will have large impacts for years to come over a large population.  An example of such decisions are sustainable development policies. The fundamental question is, how can we take actions today that align with long-term goals? Researchers in the field of robust decision-making have dealt with this question by enumerating multiple states of the world without ranking their likelihood \cite{Kwakkel2016CopingUncertainty}  States of the world is a central concept in decision theory which refers to a feature of the world that the DM has no control over and is the origin of the DM's uncertainty about the world. Each of the possibilities of the future is called \textit{scenario} and from the multi-objective problem point of view, the goal is to test the set of optimal solutions on robustness. This is usually done by checking how sensitive they are to different states of the world \cite{Robustness-McPhail-2018}. 

Kwakkel {\etal} \shortcite{Exploratory-Kwakkel-2017} propose EMA Workbench, an open-source Python library, to assist multi-objective decision making under deep uncertainty. It supports visual analysis integrated with robustness analysis \cite{Impact-McPhail-2020}, which consists of four components: (i) generation of policy options (through MOO); (ii) generation of states of the world (scenarios against which candidate policy options are evaluated); (iii) vulnerability analysis (which aims to identify the relative influence of the various uncertain factors on policy robustness); and (iv) robustness evaluation (through calculation of different metrics of robustness from the literature such as satisficing metrics, regret metrics, and descriptive statistics of the distribution of outcomes over the states of the world).

Shavazipour {\etal} \shortcite{shavazipour2021multi} propose two novel visualizations tools for scenario-based MOO, which support the DM in exploring, evaluating, and comparing the performances of different solutions according to all objectives in all plausible scenarios. These visualization methods are (i) a novel extension of empirical attainment functions for scenarios (SB-EAF); and (ii) an adapted version of heatmaps. With SB-EAF, practical visualization is limited to bi-objective optimization problems as it can become non-intuitive for a DM to analyse a large number of solutions through SB-EAF.

\section{Emerging Research Directions}
\label{sec:directions}

We reviewed three established lines of research on decision support for MOO in the previous section. As this topic is gaining traction, there are several emerging lines of research that we discuss in this section.

\paragraph{Interactive visualizations} There is an increasing emphasis on making MOO visualizations interactive. The interactivity enables a DM to, e.g., select and focus on specific solutions or subset of solutions, hide certain dimensions, or cluster solutions. Recent MOO visualization libraries such as PAVED \cite{PAVED-Cibulski-2020} and Parasol \cite{Parasol-Raseman-2019} offer such interactive features. However, there are several avenues to improve the MOO visualizations.

First, although recent tools enable interactivity, they do not guide the DM on what to visualize. For example, a DM can be guided on interesting solutions or clusters. This can be facilitated by systematically integrating data mining (Section~\ref{sec:mining_the_solution_set}) and visualization techniques (Section~\ref{sec:visualization}). Further, although most visualizations (and DM's preferences) are in the objective space, the knowledge required to implement the preferred solutions is in the decision space. Thus, there is a need for techniques that bridge the two spaces. To this end, in a recent work, Smedberg and Bandaru \shortcite{Interactive-Smedberg-2022} develop an interactive decision-support system that integrates knowledge discovery and visualization techniques. However, the effectiveness of this tool in real-world applications remains to be studied.

Second, a DM often needs to navigate multiple types of visualizations to gain a good understanding of the solution sets. Thus, it is important to facilitate the DM's knowledge discovery in an incremental manner. For instance, if the DM focused on one cluster in a plot, it should be easy to explore solutions from that cluster in another plot. Such continuity is largely missing in the current tools. Further, there is also a need to recognize the decision-making `styles' of individual DMs and personalize decision support tools, accordingly.

Finally, most of the existing MOO visualizations are meant to be used by experts (e.g., researchers and engineers). However, the DMs, e.g., a city council, may not have the MOO expertise. How to support such DMs remains a largely open question. To this end, methodologies such as data storytelling \cite{Patterns-Ojo-2018} can be beneficial, but they need to be adapted for MOO outputs and workflows.

\paragraph{State-of-the-art ML methods}
First, the existing machine learning (ML) methods for mining the solution sets are typically fully supervised or unsupervised. However, there are intermediate paradigms such as active and semi-supervised learning. Along this direction, Zintgraf {\etal} \shortcite{Zintgraf-2018-AAMAS-MODMPreference} develop a method to uncover implicit user preferences via Gaussian processes and active learning. Similar methods can be adopted for other tasks related to MOO knowledge discovery. Second, existing classification models that treat the (discretized) objective value as the class variable, consider one objective at a time or an aggregation of objective values. An unexplored direction is to consider multiple objectives at the same time via multi-label classification models, which learn from label correlations. Finally, simple data mining methods such as decision trees are preferred over black-box models such as neural networks in the knowledge discovery phase since the goal is to generate knowledge for humans (DMs). However, there have been significant advances in making neural networks explainable. Such techniques are yet to be explored for analyzing an MOO dataset (Section~\ref{sec:mining_the_solution_set}).

\paragraph{Explainable MOO} Explainability is an important topic across AI subfields. MOO methods, which produce a solution set, can possibly offer more information to the user compared to other AI methods, which produce a single solution. However, turning the trade-offs implicit in an MOO solution set into explicit explanations is largely an open challenge. It leads to important questions such as: What kinds of explanations can MOO produce? How informative are they? How easy are they for DMs to understand? How do they influence the decision-making process (positively or negatively)? First, there is a need to come up with such questions systematically. In this direction, one can build on the recent explainable AI question bank \cite{Questioning-Liao-2020} and adapt it for MOO.

There are some recent works on explainable MOO methods. Misitano {\etal} \shortcite{Towards-Misitano-2022} propose an explainability framework for interactive MOO (specifically for algorithms employing scalarization). This framework utilizes SHAP, a popular explainable AI method, and produces rules explaining the trade-offs during the optimization process. Corrente {\etal} \shortcite{Explainable-Corrente-2021} propose a similar approach for multi-objective evolutionary algorithms. However, research in this direction is still at its infancy. There are several open questions about, e.g., explaining the search process, identifying the objectives or decision variables that influence the algorithm the most, and identifying potential biases in the solution set.

\paragraph{Dynamic uncertainty and expert elicitation} The challenge of coping with epistemic uncertainty in multi-objective decision support has been largely addressed by the field of decision making under deep uncertainty.  The methods to cope with deep uncertainty described in Section~\ref{sec:uncertainty_exploration} entail a comprehensive set of tools to analyze the possible outcomes and options to reduce decision risk. Nonetheless, the majority of these approaches focus on the robustness of solutions and less so on their adaptability and flexibility. Much of the uncertainty consideration in MOO is static---it is integrated at specific stages of the analysis with little room to adapt the search process when new information is available. The interplay between dynamic uncertainty,  multi-objective optimization and decision support is widely understudied.

A key research direction is to enable feedback mechanisms between uncertainty exploration and the search space to integrate new knowledge. Such a mechanism can directly contribute to real-time robust decision support. Further, there is a lack of consensus about the importance of deeply uncertain parameters and their lower and upper bounds. To address this challenge, future research can focus on developing techniques to systematically integrate expert knowledge and reach a consensus on deeply uncertain parameters. This represents a key step towards the acceptability of solutions by grounding the uncertainty exploration on recognized expert criteria. 

\paragraph{Time sensitivity} There has been little work focusing on the analysis of the behaviour of the solutions and their trade-offs over time, with only Quinn {\etal} \shortcite{What-Quinn-2019} proposing a time-varying sensitivity analysis to evaluate how the sets of water release policies adapt and coordinate information use across the reservoirs differently. They utilise variance decomposition (local, derivative-based sensitivity analysis) of the prescribed release policies to analyse how they use state information in different ways over the course of 1000 years of simulation period. The decomposition is done for each day, for each reservoir (four reservoirs included in the case study) to determine which information source influences the release decision the most, for each reservoir during different times of the year, and across the years. Although the behaviour of the solutions over time is an important angle of the multi-objective analysis for sequential decision-making problems, it is still an understudied field of MOO.  

\paragraph{Alternative ethical framings} The fair distribution of benefits and risks among stakeholders is a major concern in decision support. MOO has partially overcome unequal distributional outcomes due to its ability to avoid aggregation over objectives into, e.g., a single economic metric. There remains, however, the opportunity to explicitly include alternative justice representations within MOO. Pareto optimality, while argued by some to be a necessary condition for justice, has been criticized for being too ideal and not ensuring fairness or stability, i.e., the focus is placed on seeking ideal solutions instead of acceptability and achievability. 

A relevant direction in current MOO research is understanding how different ethical principles and values can be integrated into multi-objective decision support systems, and how they can affect the distribution of outcomes. Ethical considerations within MOO have so far been addressed by setting performance thresholds on the solution sets, or by integrating additional fairness goals in the problem formulation.  However, the question of how to restructure the design of MOO algorithms using alternative ethical theories to guide the search is yet to be explored. This question can be approached by linking MOO design with two major branches of ethical theory---consequentialist and deontological. The first represents the status-quo and focuses on the outcomes of the solution, and the second, focuses on their adherence to a given moral rule regardless of their outcomes. Further, to understand how consensus between DMs is affected by different justice principles, decision support should provide the ability to choose between ethical theories. This will broaden our understanding of how to integrate ethical concerns in decision support systems, and provide practical insights for the design and deployment of MOO systems.

\paragraph{Understanding the DM's needs} Despite a growing body of work on decision support for multi-objective decision-making, there is a dearth of systematic studies on who the DMs are, what kind of support they need, and at what stage of the decision-making process. Much of the current work relies on the researchers' assumptions or, at best, anecdotal evidence on what the DMs need. Thus, an important research direction is to systematically understand the DMs requirements. Along this direction, for instance, Dy {\etal} \shortcite{Improving-Dy-2022} conduct an empirical study to compare the four popular visualizations (Section~\ref{sec:visualization}). In particular, they analyze how the chart complexity, the data volume (number of options and dimensions shown), and the DMs' prior experience affects the time and accuracy of decision-making. Such empirical studies are necessary, not only for visualization, but also for each aspect of decision support such as knowledge discovery methods, explainability, and uncertainty handling.

\paragraph{A reference architecture for tool development} There exist several software tools---web applications, libraries, and frameworks---for decision support on MOO (we refer to only a few of these tools in this paper). Many of these tools are open source. Further, there is a substantial overlap in features among these tools. Thus, a valuable direction (e.g., for another survey) is to systematically catalog these tools and their functionalities. The dimensions of decision support we identify (in Sections~\ref{sec:decision-support-methods} and~\ref{sec:directions}) can provide an initial structure for such a catalog. Further, at a technical level, this initial structure can be refined into a reference architecture for MOO decision-support systems. Such an architecture (with an associated inventory of tools for each component of the architecture) can both reduce the entry barrier for practice and increase the pace of innovation on MOO decision support.

\section{Conclusions}
\label{sec:conclusions}

There is growing recognition that AI systems are intended to augment (not replace) human intelligence. The MOO methods fit in this line of thought very well as they offer trade-offs DMs can explore in order to come up with final solutions as opposed to simply adopting the solutions AI suggests. Such a human-aligned decision-making process is particularly important for addressing safety and other ethical concerns of using AI, as well as legal requirements \cite{Vamplew-2018-EIT-HumanAI}.

Although MOO is a well-established topic, extant research on this topic is largely focused on the algorithmic aspects of optimization. However, the human-centeredness of the multi-objective decision-making process is increasingly recognized. Accordingly, there is a growing body of work on engaging humans in the complex decision-making process. This body of work is spread across research fields, including AI, Operations Research, and application areas like Environmental Sciences. Our work brings these works together under the umbrella of decision support methods for MOO.

We identify three categories of decision-support methods---visualization, knowledge discovery, and uncertainty exploration---that have been well-studied in the literature. We do not provide an exhaustive list of works in these categories, but provide a comprehensive overview. Importantly, we identify, a number of emerging research lines on this topic, including, interactive visualization, explainability, and support on ethical aspects. We provide concrete research directions along these lines. Finally, we also identify the needs for qualitative research on this topic, specifically to identify the needs of DMs, and call for collaborative effort to bring together practical tools.

\section*{Acknowledgements}
This research was supported by funding from the TU Delft AI Initiative and the Flemish Government under the ``Onderzoeksprogramma Artifici\"{e}le Intelligentie (AI) Vlaanderen'' program. We would also like to thank Yasin Sari for providing data and code to generate visualizations in Figure~\ref{fig:moo-visuals}.

\bibliographystyle{named}

 \providecommand{\url}[1]{#1}

\end{document}